\documentclass[11pt,a4paper]{article}

\usepackage[T1]{fontenc}
\usepackage[utf8]{inputenc}
\usepackage[a4paper,margin=2.3cm]{geometry}
\usepackage{graphicx}
\usepackage{multirow}
\usepackage{amsmath,amssymb,amsfonts}
\usepackage{booktabs}
\usepackage{tabularx}
\usepackage{array}
\usepackage{algorithm}
\usepackage{algorithmic}
\usepackage{subcaption}
\usepackage{xspace}
\usepackage{microtype}
\usepackage{placeins}
\usepackage{float}
\usepackage[numbers,sort&compress]{natbib}
\usepackage[hidelinks]{hyperref}

\newcommand{\dBC}{Diffusion-BC\xspace}
\newcommand{\dTwoBC}{Diffusion-2BC\xspace}
\newcommand{\mseBC}{MSE-BC\xspace}

\title{Diffusion-2BC: Hybrid Diffusion and Regression Training\\
for Multimodal Offline Behavior Cloning in Autonomous Driving}
\author{Bruno Maciel Machado \\ Eric Aislan Antonelo\\[0.5em]
\small Department of Automation and Systems Engineering\\
\small Graduate Program in Automation and Systems Engineering\\
\small Federal University of Santa Catarina, Florian\'opolis, Santa Catarina, Brazil\\
\small Corresponding author: \href{mailto:eric.antonelo@ufsc.br}{eric.antonelo@ufsc.br}}
\date{}

\begin{document}
\maketitle

\begin{abstract}
Behavior cloning (BC) offers a simple offline route to learning policies from demonstrations, but conventional mean-squared-error regression is poorly suited to datasets in which similar observations admit several valid actions. We investigate whether a direct regression objective can improve the training of diffusion-based behavior cloning while retaining stochastic multimodal inference. We propose Diffusion-2BC, which combines a diffusion denoising objective with an auxiliary deterministic BC loss over a shared visual encoder; the auxiliary branch is removed at test time, so inference remains diffusion-based. The method is evaluated in the Claw multimodal benchmark and in bird's-eye-view CARLA navigation, including route-conditioned driving, route-free multi-intersection navigation, and descriptive cross-map evaluation from Town01 to Town02. In Claw, Diffusion-2BC reduces mean mask-distance error by approximately 10\% relative to Diffusion-BC and by 85\% relative to deterministic BC. In route-conditioned CARLA, it reaches infraction-free completion earlier and with lower variability than Diffusion-BC. In the route-free experiment, it attains the highest observed mean distance under the configured evaluation protocol, while selected rollouts provide qualitative evidence of distinct route choices. Within the evaluated datasets and baselines, these results show that an auxiliary regression signal can improve the closed-loop reliability of diffusion behavior cloning while retaining multimodal prediction.
\end{abstract}

\vspace{1em}

\section{Introduction}
Behavior cloning (BC) provides a simple offline approach to policy learning from demonstrations, but conventional regression objectives are poorly suited to datasets in which similar observations admit multiple valid actions. This issue is particularly relevant in autonomous driving, where different steering or route choices may all be compatible with a given local observation. At an intersection, for example, a driver may turn left or right; when approaching a bend, different combinations of steering and acceleration may remain safe. Such alternatives are not necessarily annotation errors. They can be valid modes of expert behavior and should be represented by a policy intended to imitate diverse demonstrations.

BC treats policy learning as supervised prediction from observations to actions \cite{Pomerleau1989Alvinn,Imitation_Learning_A_Survey_of_Learning_Methods}. Its simplicity makes it computationally efficient and fully compatible with fixed offline datasets. Nevertheless, two limitations are important in closed-loop use. First, errors compound because the learned policy encounters states that were not represented in the expert data \cite{Efficient_Reductions_for_Imitation_Learning,Imitation_Not_Enough_Robustifying_Imitation}. Second, the usual mean-squared-error (MSE) objective represents the conditional mean of the action distribution. When demonstrations contain separated modes, that mean may correspond to an undesirable action or the network may collapse to a single dominant mode \cite{Implicit_Behavioral_Cloning,Behavior_Transformers_Cloning_k_modes_with_one_stone}.

Generative policies offer a way to model a distribution rather than a point estimate. Diffusion models learn to reverse a progressive noising process and have been effective in high-dimensional generation \cite{Deep_Unsupervised_Learning_using_Nonequilibrium_Thermodynamics,Denoising_Diffusion_Probabilistic_Models,Improved_Denoising_Diffusion_Probabilistic_Models}. Recent work has adapted them to planning, decision-making, and visuomotor control \cite{janner2022planningdiffusionflexiblebehavior,Conditional_Generative_Modeling_for_Decision_Making,Diffusion_Policy_Visuomotor_Policy_Learning,Imitating_Human_Behaviour_with_Diffusion_Models}. In behavior cloning, the reverse diffusion process can generate different actions for similar observations, which directly addresses multimodality without discretizing continuous controls.

Expressiveness alone, however, does not guarantee reliable closed-loop driving. A diffusion policy must learn useful visual features and a denoising process from a finite expert dataset, and every action requires multiple network evaluations. Our experiments indicated that a basic diffusion behavior-cloning policy could express conflicting left/right decisions but could also be inconsistent in complex route-free urban navigation. This motivates the central question of this paper: \emph{can a direct deterministic behavior-cloning signal improve the representation learning and closed-loop performance of a diffusion policy without eliminating multimodal action generation?}

We answer this question with \dTwoBC, a hybrid training architecture. A convolutional feature extractor feeds both a diffusion denoising branch and an auxiliary fully connected action-regression branch. Their losses are combined by a coefficient $\alpha$. The auxiliary branch is present only during training, so the test-time policy remains stochastic and uses the same reverse process as standard diffusion behavior cloning. The method is evaluated entirely offline: no environment interaction is used to update the policy after the demonstration datasets have been collected.

The evaluation is organized as a focused experimental study. The Claw benchmark tests multimodal conditional prediction, whereas CARLA provides the main closed-loop evaluation of route following, route-free navigation, qualitative route variation, and cross-map behavior. This organization makes explicit which conclusions are supported by each environment and keeps the preliminary experiments separate from the validation of the proposed method.

\subsection{Relation to preliminary work}

A preliminary conference paper evaluated standard diffusion behavior cloning against deterministic behavior cloning in CarRacing and CARLA,

\cite{Machado2025CROS}. The present work substantially extends that study by introducing the \dTwoBC{} architecture and hybrid objective, evaluating the proposed method in route-conditioned and route-free multi-intersection CARLA navigation, adding the Town01-to-Town02 cross-map analysis, and expanding the methodological and computational-cost

discussions.

Before developing \dTwoBC{}, the conference study compared deterministic \mseBC{} and \dBC{} in CarRacing \cite{Machado2025CROS}, providing initial evidence that diffusion-based action generation can improve visual closed-loop control. This result motivated the addition of a direct regression signal to diffusion training. Supplementary Note~S1 summarizes the experiment and provides additional visualizations.

The contributions are as follows:
\begin{itemize}
    \item We formulate \dTwoBC, an offline behavior-cloning architecture that combines denoising and deterministic regression objectives in a shared visual representation while preserving diffusion-only inference.
    \item We compare \mseBC, \dBC, and \dTwoBC under matched experimental conditions in the Claw benchmark and in route-conditioned and route-free CARLA navigation, including a descriptive cross-map evaluation from Town01 to Town02.
    \item We use this focused comparison to isolate the effect of the auxiliary regression signal and analyze route progress, multimodal behavior and computational cost.

\end{itemize}

\section{Related Work and Positioning}

\subsection{Behavior cloning for autonomous driving}
Early end-to-end driving systems already used supervised imitation to map visual input to steering commands \cite{Pomerleau1989Alvinn}. Modern behavior-cloning systems use deeper visual encoders and richer sensor representations, but the central supervised formulation remains unchanged. Conditional imitation learning addresses route ambiguity by providing a high-level command such as turn left, turn right, or continue straight \cite{End_to_end_Driving_via_Conditional_Imitation_Learning}. Such conditioning is effective when command labels are available, but it does not solve the distinct problem considered here: learning the distribution of unlabeled valid actions that are jointly present in the demonstrations.

Driving-policy errors also arise from distribution shift. Small mistakes move the vehicle away from expert states, after which predictions can degrade rapidly. Dataset aggregation and robustification methods reduce this problem by collecting additional states or synthesizing perturbations \cite{Efficient_Reductions_for_Imitation_Learning,Imitation_Not_Enough_Robustifying_Imitation,ChauffeurNet_Learning_to_Drive_by_Imitating_the_Best_and_Synthesizing_the_Worst}. The present study does not use online relabeling or additional recovery demonstrations; it asks how far an offline policy can progress using the available fixed datasets.

Learning-based and hybrid controllers have also been evaluated for autonomous driving in CARLA. Hiremath and Bajcinca \cite{Hiremath2026Lifelong} combine a learned controller with model-based control and imitation-based adaptation for lifelong path following. Their setting emphasizes hybrid online adaptation, whereas the present work focuses on fully offline imitation from fixed demonstrations and on representing multimodal action choices without online policy updates.

\subsection{Multimodal imitation learning}
Several approaches have been proposed for multimodal action distributions. Discretization converts continuous control into classification but introduces quantization and can separate dimensions that should remain coordinated. Mixture-density and latent-variable models represent several modes explicitly, while energy-based methods score candidate actions without requiring a normalized output distribution. Implicit Behavioral Cloning, for example, models the observation-action compatibility through an energy function \cite{Implicit_Behavioral_Cloning}. Behavior Transformers represent multiple action modes using discretized latent structure \cite{Behavior_Transformers_Cloning_k_modes_with_one_stone}. These methods demonstrate that a unimodal regression loss is not the only viable BC objective.

The experimental comparison is designed to isolate the effect of the auxiliary regression signal rather than to provide an exhaustive benchmark of multimodal imitation-learning methods. \mseBC provides the deterministic reference, \dBC provides the same diffusion-policy family without the auxiliary regression signal, and \dTwoBC adds that signal while retaining the same diffusion inference mechanism. The comparison therefore establishes the effect of the proposed training modification relative to these matched baselines, not superiority over every multimodal imitation-learning family.

\subsection{Diffusion models for decision-making and control}
Diffusion models define a forward Markov chain that progressively corrupts data and a learned reverse chain that reconstructs samples \cite{Denoising_Diffusion_Probabilistic_Models}. Conditional diffusion extends this mechanism by conditioning the denoising network on context. In decision-making, diffusion models have been used to generate trajectories, action sequences, and control outputs \cite{Conditional_Generative_Modeling_for_Decision_Making,Diffusion_Model_Effective_Planner_Multi_Task_RL,Diffusion_Policy_Visuomotor_Policy_Learning}. Their iterative sampling supports multimodality but increases inference cost.

The closest baseline in this work, \dBC, conditions denoising on features extracted from the current observation and predicts the noise applied to an expert action. \dTwoBC adds an auxiliary deterministic head and MSE loss during training. This resembles the broader principle of supplementing a difficult imitation objective with a direct BC signal, as explored in BC-augmented adversarial imitation learning \cite{Augmenting_GAIL_with_BC_for_Sample_Efficient_Imitation_Learning}. The distinction is that \dTwoBC remains fully offline, does not train a discriminator, and removes the deterministic branch at inference.

Diffusion Model-Augmented Behavioral Cloning also combines a BC objective with a diffusion-model objective, but it learns a conventional conditional policy together with a diffusion model of the joint expert state--action distribution \cite{Diffusion_Model_Augmented_Behavioral_Cloning}. In contrast, \dTwoBC{} uses the conditional diffusion model itself as the executed policy and employs deterministic regression only as an auxiliary training signal. Thus, the methods share the idea of complementary conditional and diffusion losses but assign the diffusion component different roles at inference.

\begin{table}[!htbp]
\centering
\caption{Qualitative positioning of the methods most directly related to this study. ``Unlabeled multimodality'' means that several valid actions may appear for similar observations without an explicit command identifying the desired mode.}
\label{tab:positioning}
\scriptsize
\setlength{\tabcolsep}{3.2pt}
\begin{tabularx}{\textwidth}{>{\raggedright\arraybackslash}p{0.17\textwidth} >{\raggedright\arraybackslash}X >{\raggedright\arraybackslash}p{0.22\textwidth} >{\raggedright\arraybackslash}p{0.27\textwidth}}
\toprule
Method family & Representation / multimodality & Main strength & Limitation relative to this study \\
\midrule
MSE behavior cloning & Deterministic point estimate; no unlabeled multimodality & Simple, fast offline training and inference & Conditional averaging or single-mode behavior \\
Command-conditioned BC & Deterministic action plus route command; modes require labels & Direct controllability at intersections & Requires reliable command or route labels \\
Energy-based / implicit BC & Energy over observation--action pairs; supports unlabeled modes & Flexible multimodal scoring & Requires candidate optimization or sampling; not evaluated here \\
Behavior Transformer & Discrete latent/action representation; supports several modes & Represents several behavior modes & Quantization and architecture choices differ from continuous diffusion \\
Diffusion-augmented BC & Conditional BC policy plus joint state--action diffusion model & Combines conditional and joint expert modeling & The direct policy, rather than the diffusion model, is the primary action predictor \\
Diffusion-BC & Conditional generative action model; supports unlabeled modes & Continuous stochastic action generation & Iterative inference and inconsistent closed-loop behavior in the hardest experiment \\
Diffusion-2BC (proposed) & Diffusion policy with auxiliary regression training signal & Preserves stochastic inference while improving shared features & Additional training complexity; not compared with all multimodal BC families \\
\bottomrule
\end{tabularx}
\end{table}

\section{Background}
\subsection{Offline behavior cloning}
Let $\mathcal{D}=\{(o_i,a_i)\}_{i=1}^{N}$ denote a fixed dataset of expert observation-action pairs. The deterministic baseline learns a policy $\pi_\theta$ by minimizing
\begin{equation}
\mathcal{L}_{\mathrm{MSE}}=
\mathbb{E}_{(o,a)\sim\mathcal{D}}
\left[\|\pi_\theta(o)-a\|_2^2\right].
\label{eq:mse}
\end{equation}
For a conditional action distribution with one concentrated mode, this objective is appropriate. For separated modes, its minimizer is the conditional mean, which need not coincide with a demonstrated action. In practice, a finite neural network may instead converge to one mode. Both outcomes are undesirable when the goal is to reproduce diverse valid strategies.

\subsection{Forward diffusion process}
Let $a_0$ be an expert action. A variance schedule $\{\beta_t\}_{t=1}^{T}$ defines $\alpha_t=1-\beta_t$ and $\bar{\alpha}_t=\prod_{s=1}^{t}\alpha_s$. The forward process adds Gaussian noise:
\begin{equation}
q(a_t\mid a_{t-1})=\mathcal{N}\left(\sqrt{\alpha_t}a_{t-1},\beta_t I\right).
\end{equation}
Using the reparameterized closed form, a noisy action at any timestep can be sampled directly:
\begin{equation}
a_t=\sqrt{\bar{\alpha}_t}a_0+
\sqrt{1-\bar{\alpha}_t}\epsilon,
\qquad \epsilon\sim\mathcal{N}(0,I).
\label{eq:forward}
\end{equation}
As $t$ increases, the action becomes progressively less distinguishable from Gaussian noise.

\subsection{Conditional reverse process for behavior cloning}
A denoising network $\epsilon_\theta$ predicts the noise used in Eq.~\eqref{eq:forward}. It receives the noisy action $a_t$, the diffusion timestep $t$, and features $f_\theta(o)$ extracted from the current observation. The simplified training objective is
\begin{equation}
\mathcal{L}_{\mathrm{DBC}}=
\mathbb{E}_{o,a,t,\epsilon}
\left[\|\epsilon_\theta(f_\theta(o),a_t,t)-\epsilon\|_2^2\right].
\label{eq:dbc}
\end{equation}
At inference, the policy samples $a_T\sim\mathcal{N}(0,I)$ and repeatedly applies the learned reverse update until an action estimate $a_0$ is obtained. Different initial noise samples can produce different valid actions under the same observation, which is the mechanism used to express multimodality.

\section{Proposed Diffusion-2BC Method}
\subsection{Design rationale}
The experiments motivating the method showed complementary behavior. Deterministic BC was fast and could learn route following but failed to represent conflicting decisions at the same intersection. Diffusion BC represented left and right turns with a single model but was less consistent in the more difficult route-free, multi-intersection setting. We therefore retain diffusion inference and add a direct regression signal only during representation learning.

\subsection{Architecture and objective}
\begin{figure}[t]
\centering
\includegraphics[width=0.9\columnwidth]{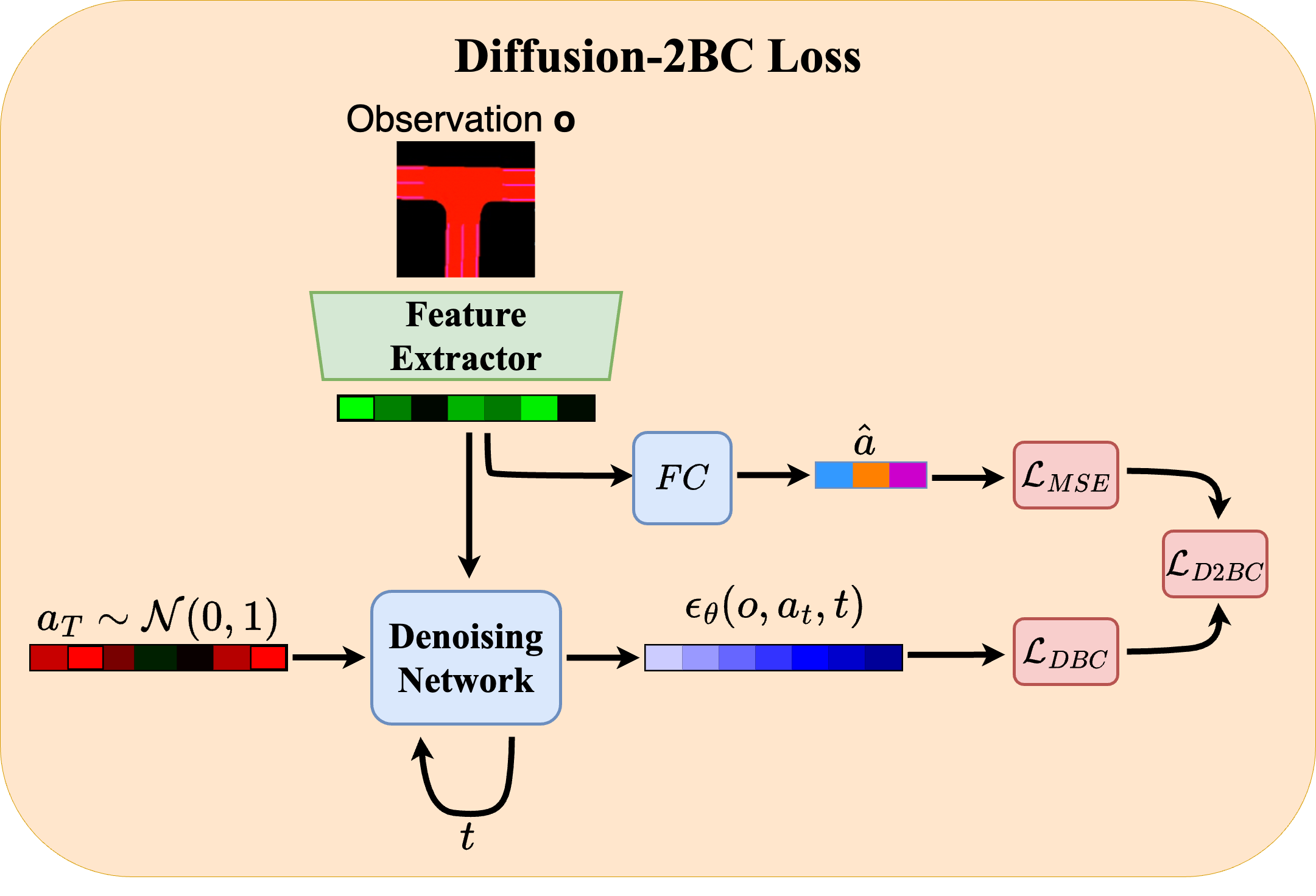}
\caption{Diffusion-2BC training architecture. The shared feature extractor processes the observation once. Its representation conditions the denoising branch used to compute $\mathcal{L}_{\mathrm{DBC}}$ and feeds an auxiliary deterministic head used to compute $\mathcal{L}_{\mathrm{MSE}}$. Their weighted sum updates the shared representation. The auxiliary head is discarded after training, and test-time actions are generated only by reverse diffusion}
\label{fig:d2bc}
\end{figure}

Fig.~\ref{fig:d2bc} shows the training architecture. The observation is first processed by a shared convolutional feature extractor $f_\theta$. Its output conditions a denoising network and is also passed to an auxiliary fully connected head $g_\theta$ that predicts the expert action directly. The combined objective is
\begin{equation}
\mathcal{L}_{\mathrm{D2BC}}=
\alpha\mathcal{L}_{\mathrm{MSE}}+(1-\alpha)\mathcal{L}_{\mathrm{DBC}},
\qquad 0\leq\alpha\leq1,
\label{eq:d2bc}
\end{equation}
where
\begin{equation}
\mathcal{L}_{\mathrm{MSE}}=
\|g_\theta(f_\theta(o))-a\|_2^2.
\end{equation}
The coefficient $\alpha$ can be fixed or scheduled. A fixed value of 0.3 is used in the route-conditioned CARLA experiment. In the multi-intersection experiment, $\alpha$ is updated once per epoch using $\alpha_e=\exp(-k e)$, with $k=-\ln(0.01)/E$, so that the schedule starts at 1.0 and decays toward 0.01 over $E$ training epochs. The study does not include a systematic hyperparameter ablation, so these settings are treated as practical design choices rather than globally optimal values.

\subsection{Network architecture}
All evaluated policies use the same visual encoder. It comprises two residual convolutional blocks, each with two $3\times3$ convolutions (stride 1, padding 1), batch normalization, and GELU activations. The blocks use 64 and 128 output channels, respectively; their residual sums are scaled by $1/\sqrt{2}$, and each block is followed by $2\times2$ max pooling. An $8\times8$ average-pooling layer and flattening produce the visual representation. The input and final action projections are adapted to each environment.

The conditional denoising backbone follows the observation-to-action Diffusion-BC architecture of Pearce et al.~\cite{Imitating_Human_Behaviour_with_Diffusion_Models}. The visual representation, noisy action, and normalized diffusion timestep are embedded separately into 128-dimensional vectors and projected to three 64-dimensional tokens. Four Transformer encoder blocks process these tokens, each using 16 attention heads, residual connections, batch normalization, and a feed-forward expansion from 64 to 256 units with GELU activation. The flattened Transformer output is projected to the action-noise dimension. No Transformer dropout is used in the reported CARLA runs, and classifier-free context dropout and guidance are disabled ($p_{\mathrm{ctx}}=0$, guidance weight $=0$).

For CARLA, the convolutional encoder produces a 4,608-dimensional flattened representation. The deterministic \mseBC{} baseline maps this vector through hidden widths 2,304, 1,024, and 64 to the two-dimensional control output. \dTwoBC{} retains the same denoising branch as \dBC{} and adds an auxiliary regression head $4608\rightarrow2048\rightarrow512\rightarrow128\rightarrow32\rightarrow2$, with BatchNorm, ReLU, and dropout $p=0.3$ after each hidden layer. The auxiliary head is used only during training. Table~\ref{tab:architecture} summarizes the resulting policy structures, while Supplementary Table S2 provides the layer-by-layer specification.

\begin{table}[t]
\centering
\caption{Compact comparison of the evaluated policy architectures}
\label{tab:architecture}
\scriptsize
\begin{tabularx}{\columnwidth}{>{\raggedright\arraybackslash}p{0.20\columnwidth}XXX}
\toprule
Component & \mseBC{} & \dBC{} & \dTwoBC{} \\
\midrule
Visual encoder & Residual CNN & Residual CNN & Shared residual CNN \\
Executed head & CARLA: MLP $4608$--$2304$--$1024$--$64$--$2$ & Transformer denoiser & Transformer denoiser \\
Auxiliary head & -- & -- & CARLA: MLP $4608$--$2048$--$512$--$128$--$32$--$2$ \\
Inference & One deterministic pass & $T$ reverse steps & $T$ reverse steps \\
\bottomrule
\end{tabularx}
\end{table}

\subsection{Training and inference}
Algorithm~\ref{alg:d2bc} summarizes vectorized minibatch training. A diffusion timestep and Gaussian noise are sampled independently for every action in the minibatch. The corrupted actions and observation features are processed by the denoising branch, while the same features are processed by the auxiliary regression branch. The two minibatch-averaged losses are combined before a single parameter update. During inference, the auxiliary prediction and regression loss are absent; consequently, \dTwoBC has the same test-time computational structure as \dBC.

\begin{algorithm}[t]
\caption{Diffusion-2BC minibatch training}
\label{alg:d2bc}
\begin{algorithmic}[1]
\REQUIRE Demonstrations $\mathcal{D}$; timesteps $T$; diffusion schedule $\bar\alpha_{1:T}$; loss-weight schedule $\alpha_e$
\FOR{each training epoch $e$}
  \STATE Set $\alpha\leftarrow\alpha_e$ once for the current epoch
  \FOR{each minibatch $(O,A)\sim\mathcal{D}$ of size $B$}
    \STATE Sample $t_i\sim\mathcal{U}\{1,\ldots,T\}$ and $\epsilon_i\sim\mathcal{N}(0,I)$ for $i=1,\ldots,B$
    \STATE $A_t\leftarrow\sqrt{\bar\alpha_t}\odot A+\sqrt{1-\bar\alpha_t}\odot\epsilon$
    \STATE $H\leftarrow f_\theta(O)$
    \STATE $\hat\epsilon\leftarrow\epsilon_\theta(H,A_t,t)$; $\hat A\leftarrow g_\theta(H)$
    \STATE $L_{\mathrm{DBC}}\leftarrow B^{-1}\sum_{i=1}^{B}\|\hat\epsilon_i-\epsilon_i\|_2^2$
    \STATE $L_{\mathrm{MSE}}\leftarrow B^{-1}\sum_{i=1}^{B}\|\hat A_i-A_i\|_2^2$
    \STATE $L\leftarrow(1-\alpha)L_{\mathrm{DBC}}+\alpha L_{\mathrm{MSE}}$
    \STATE Update all trainable parameters using $\nabla L$
  \ENDFOR
\ENDFOR
\end{algorithmic}
\end{algorithm}

\subsection{Interpretation of the auxiliary loss}
The auxiliary head should not be interpreted as a second policy used to arbitrate with the diffusion output. Its role is representation learning. The direct action target provides a short path from the expert controls to the visual features, while the denoising branch retains the distributional objective. This separation explains why a nonzero MSE weight need not force deterministic inference: stochasticity is preserved because the deterministic prediction is not executed.

The formulation also clarifies two boundary cases. With $\alpha=0$, the model reduces to \dBC. As $\alpha$ approaches one, denoising receives progressively less weight and the training objective approaches deterministic BC, although the remaining reverse model would be poorly trained. Useful operation therefore requires both terms. The experiments therefore evaluate two practical choices---a fixed weight and an exponentially decaying schedule---but leave a full schedule ablation to future work.

\section{Experimental Methodology}
\subsection{Staged evaluation strategy}
The experiments are organized according to the research question addressed by each environment:
\begin{enumerate}
    \item \textbf{Multimodal prediction:} Can diffusion-based BC and the hybrid objective reproduce several valid actions in the Claw environment?
    \item \textbf{Route-conditioned urban navigation:} Can the methods learn stable trajectory following in CARLA from a small route dataset?
    \item \textbf{Route-free urban navigation and transfer:} Can the methods drive through multiple intersections, express varied paths, and transfer from Town01 to unseen Town02?
\end{enumerate}
This progression prevents the strongest CARLA claims from being inferred solely from simplified environments.

\subsection{Claw environment}
We adopt the Claw environment introduced in the original Diffusion-BC study~\cite{Imitating_Human_Behaviour_with_Diffusion_Models} to evaluate multimodal behavior cloning. The environment uses $32\times32$ RGB top-down images containing one or more of five toy shapes. The shapes are randomly translated, rotated, and scaled on a fixed background. The action is a two-dimensional point $a=(x,y)$, and an action is successful when it lies inside the mask of any object in the image. One successful point is sampled for each training image. Images containing multiple objects therefore admit several disconnected valid action regions, making the task a compact test of multimodal conditional prediction.

The training set contains 20,000 observation-action pairs. The models are trained for 150 epochs with learning rate $10^{-4}$, cosine learning-rate decay, 512 hidden units, and batch size 32. Diffusion models use 50 denoising steps and a linear $\beta$ schedule from $10^{-4}$ to $2\times10^{-2}$. Evaluation uses seven held-out observations and 300 sampled actions per observation. For each sampled action, the metric is the Euclidean distance to the nearest valid mask point; predictions inside a valid object have zero error. The reported score is the mean distance over all sampled actions and observations. The same protocol is repeated after reducing the training set to 90\% of its original size.

The original Diffusion-BC paper presents representative input, mask, and sampled-action visualizations for this environment \cite{Imitating_Human_Behaviour_with_Diffusion_Models}; that figure is not reproduced here. The present study uses the same benchmark structure and reports a quantitative comparison of \mseBC{}, \dBC{}, and the proposed \dTwoBC{} objective.

\subsection{CARLA observations and controls}
\begin{figure}[t]
\centering
\begin{subfigure}{0.4\linewidth}
\centering
\includegraphics[width=\linewidth]{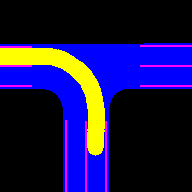}
\caption{Route-conditioned BEV}
\end{subfigure}\hfill
\begin{subfigure}{0.4\linewidth}
\centering
\includegraphics[width=\linewidth]{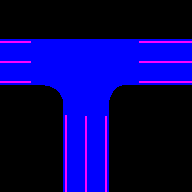}
\caption{Route-free BEV}
\end{subfigure}
\caption{Examples of the CARLA observations supplied to the policies. In (a), the yellow planned-route channel identifies the reference path, while blue and magenta encode the drivable road and lane boundaries. In (b), the route channel is omitted, so the policy must choose among locally valid road continuations. In both configurations, black denotes the non-drivable background.}
\label{fig:bevinputs}
\end{figure}

CARLA is an open-source urban-driving simulator with physics-based vehicle dynamics and configurable maps \cite{CARLA_An_Open_Urban_Driving_Simulator}. The experiments use a vehicle-centered bird's-eye-view (BEV) representation rather than raw front cameras. The BEV is a 192$\times$192 RGB image in which blue denotes the drivable road, magenta denotes lane boundaries, black denotes the non-drivable background, and yellow denotes the planned route when route conditioning is enabled. The two-dimensional action contains steering in $[-1,1]$ and an acceleration command implemented as throttle in $[0,1]$. Each BEV frame is converted to grayscale, stacked with the previous three frames to provide temporal information, and normalized, producing a $192\times192\times4$ tensor. The four frames are consecutive (frame stride 1), with no frame skipping. For the first three indices, the current frame is repeated four times to complete the input stack. The shared feature extractor uses convolutions, residual blocks, pooling, and a dense representation. \mseBC{} maps this representation directly to the action, whereas \dBC{} and \dTwoBC{} condition a transformer-based denoising network. Fig.~\ref{fig:bevinputs} shows examples of the two CARLA observation configurations.

BEV input reduces the perception burden and lets the study concentrate on policy learning. It should not be interpreted as eliminating the perception problem in a deployable system: a separate model or map-based process would be needed to construct BEV from real sensors. Previous work shows that cross-view transformations can learn this mapping \cite{Couto_2024,Santos2025,park2023cvt}.

\subsection{Route-conditioned CARLA protocol}
A PID expert follows one CARLA Leaderboard route in Town01, producing approximately 1,900 observation-action pairs. The learning agent receives road, lane-boundary, and planned-route channels but not the expert's dense control trajectory. Each architecture is trained three times. Evaluations occur every 50 epochs and use ten new random routes per model, with different lengths and curvatures. A rollout terminates at an infraction or after 3,000 steps. The reported metrics are distance traveled in meters and the fraction of the 3,000-step horizon completed.

Common hyperparameters are learning rate $10^{-4}$, fixed learning rate, 128 hidden units in the feature extractor, and batch size 32. Optimization uses Adam with zero weight decay. Diffusion models use a linear $\beta$ schedule from $10^{-4}$ to $2\times10^{-2}$, $T=20$, embedding dimension 128, and a transformer denoising architecture. \dTwoBC uses fixed $\alpha=0.3$ in this experiment.

\subsection{Route-free multi-intersection protocol}
The main multimodal dataset contains 34 PID-expert routes in Town01 and approximately 16,200 samples. It includes multiple intersection maneuvers, but the route channel is omitted from the agent's input. Two models per architecture are trained through epoch 700 and evaluated at 10-epoch intervals. Each model is evaluated five times from the same initial location. No explicit random seed is set for training. The objectives are to travel as far as possible under the active termination criteria and to determine whether repeated rollouts take distinct valid paths.

Because no reference route is active, CARLA does not provide route length. Distance is accumulated from the three-dimensional simulator coordinates:
\begin{align}
\Delta x_t &= x_t-x_{t-1}, & \Delta y_t &= y_t-y_{t-1},\\
d_t &= \sqrt{\Delta x_t^2+\Delta y_t^2}, & D &= \sum_t d_t.
\label{eq:uom}
\end{align}
Because this simulator-specific quantity is used only for within-protocol comparison and does not correspond directly to a standardized physical distance under the present setup, it is reported as a unit of measurement (UoM) rather than meters. The wrong-lane termination rule is relaxed to allow wider turns, and red-light and stop-sign termination checks are disabled in the route-free protocol; the remaining configured infractions still terminate the rollout. The same trained policies are evaluated in Town01 and in Town02, which is not used for training. For each architecture, the five checkpoints with the highest mean distance in Town01 are fixed first and then transferred to Town02 for evaluation, without a new checkpoint search in Town02.

\begin{figure}[t]
\centering
\begin{subfigure}{0.47\textwidth}
\includegraphics[width=\linewidth,angle=270]{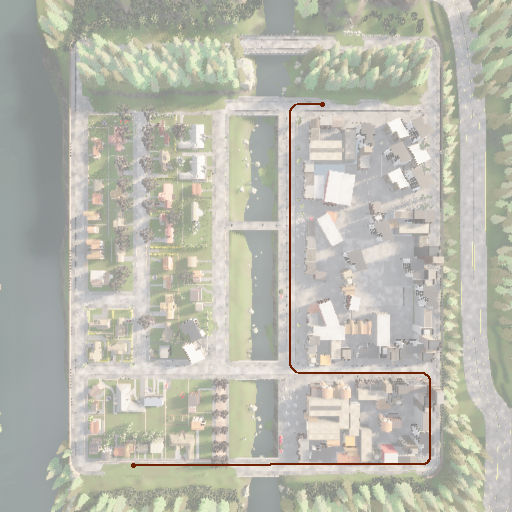}
\caption{Single route for route-conditioned training}
\end{subfigure}\hfill
\begin{subfigure}{0.47\textwidth}
\includegraphics[width=\linewidth,angle=270]{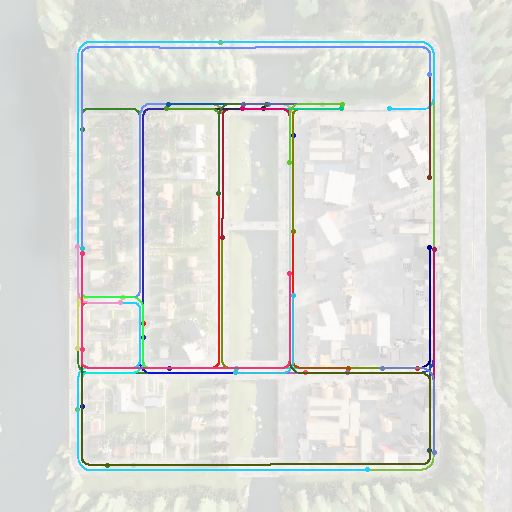}
\caption{Thirty-four routes for route-free training}
\end{subfigure}
\caption{CARLA demonstration datasets. Each colored line is a PID-expert trajectory recorded with the corresponding observations and controls. The route-conditioned dataset supplies a visible planned-route channel. The route-free dataset omits that channel and includes multiple intersection maneuvers, so similar BEV observations can be associated with different valid actions}
\label{fig:datasets}
\end{figure}

\begin{table}[t]
\centering
\caption{Summary of experimental protocols.}
\label{tab:protocols}
\scriptsize
\setlength{\tabcolsep}{3.0pt}
\begin{tabularx}{\textwidth}{>{\raggedright\arraybackslash}p{0.18\textwidth} >{\raggedright\arraybackslash}X >{\raggedright\arraybackslash}p{0.25\textwidth} >{\raggedright\arraybackslash}p{0.20\textwidth}}
\toprule
Experiment & Data and training & Evaluation and metrics & Evaluation role \\
\midrule
Claw environment & 20,000 pairs (plus a 90\% subset); \mseBC, \dBC, and \dTwoBC; simplified supervised task & 300 predictions; mean mask-distance error & Multimodal prediction \\
Route-conditioned CARLA & $\sim$1,900 pairs from one Town01 route; all three methods; three trainings with evaluation every 50 epochs & Ten new random routes; 3,000-step limit; distance and relative completion & Novel routes in the training map \\
Route-free CARLA & 34 Town01 routes ($\sim$16,200 pairs); all three methods; two trainings through epoch 700 with evaluation every 10 epochs & Five repeated rollouts from the same start; UoM distance and path diversity & Town01-selected checkpoints transferred to Town02 \\
\bottomrule
\end{tabularx}
\end{table}

For convenient replication, Supplementary Table S3 consolidates the principal training and diffusion hyperparameters used in the CARLA experiments.

\section{Results}
\subsection{Claw environment results}
\begin{table}[t]
\centering
\caption{Results in the Claw environment. The score is the mean distance to the nearest valid mask region (lower is better), averaged over seven held-out observations with 300 sampled actions per observation}
\label{tab:claw}
\small
\begin{tabular}{lcc}
\toprule
Method & Full dataset & 90\% dataset \\
\midrule
\mseBC & 4.02 & 4.25 \\
\dBC & 0.68 & 0.70 \\
\dTwoBC, exponential $\alpha$ & 0.63 & 0.69 \\
\dTwoBC, fixed $\alpha=0.3$ & \textbf{0.61} & \textbf{0.63} \\
\bottomrule
\end{tabular}
\end{table}

Table~\ref{tab:claw} reports the complete Claw results. With the full dataset, \mseBC obtains a mean distance of 4.02, whereas \dBC and the two \dTwoBC configurations lie between 0.61 and 0.68. This pattern agrees with the benchmark structure: deterministic predictions tend to concentrate, whereas diffusion samples can cover separated valid regions. With 90\% of the training data, fixed-$\alpha$ \dTwoBC reaches 0.63, compared with 0.70 for \dBC. These results show that the hybrid objective preserves the advantage of diffusion-based multimodal prediction in this benchmark; they are not used as evidence of autonomous-driving performance.

\subsection{Route-conditioned CARLA navigation}
\begin{figure}[t]
\centering
\begin{subfigure}{0.49\textwidth}
\includegraphics[width=\linewidth]{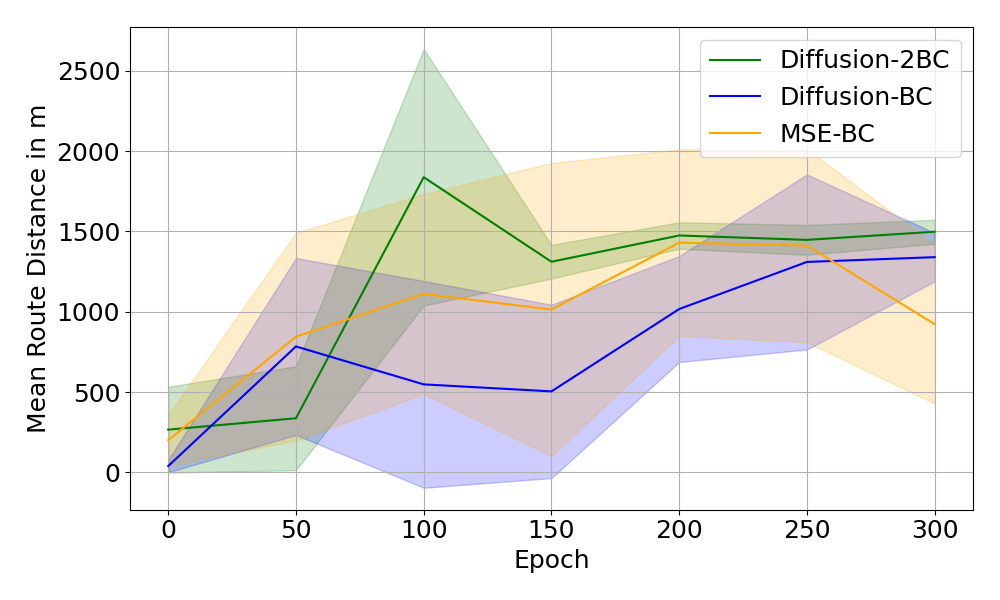}
\caption{Distance traveled before termination}
\end{subfigure}\hfill
\begin{subfigure}{0.49\textwidth}
\includegraphics[width=\linewidth]{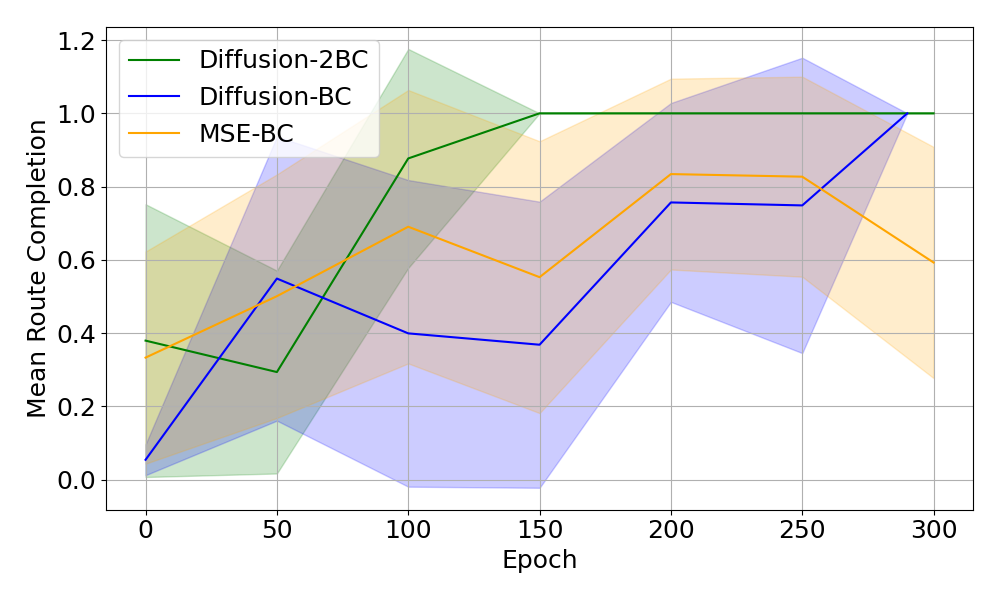}
\caption{Fraction of the 3,000-step horizon completed}
\end{subfigure}
\caption{Route-conditioned CARLA evaluation during training. Each point combines three training runs and ten random evaluation routes per model. The curves show both progress and variability; an infraction terminates the rollout before the maximum horizon}
\label{fig:fixedcurves}
\end{figure}

Fig.~\ref{fig:fixedcurves} tracks distance and completion through training. \dTwoBC is the first method to complete every evaluation rollout without an infraction, reaching that behavior at epoch 150. \dBC reaches full completion at epoch 300. \mseBC occasionally travels long distances but does not consistently complete all routes within the evaluated horizon and shows substantially higher dispersion.

At the checkpoints summarized in Table~\ref{tab:fixed}, \dTwoBC travels $1498.09\pm74.81$ m with completion $1.00\pm0.00$. \dBC also reaches complete rollouts but covers $1339.85\pm150.60$ m. The apparent possibility of a distance above or below another method despite equal completion arises because the ten randomly generated routes vary and because the vehicle speed affects distance covered within the 3,000-step horizon. The results indicate that the hybrid objective improves learning speed and stability in this route-following setting; they do not test multimodal route choice because the intended route is visible in the observation.

\begin{table}[t]
\centering
\caption{Route-conditioned CARLA evaluation at the best completion-oriented checkpoints: epoch 200 for MSE-BC, epoch 300 for Diffusion-BC, and epoch 300 for Diffusion-2BC. Values aggregate three independently trained models, each evaluated on ten random routes; the reported $\pm$ values are standard deviations over the resulting 30 rollouts}
\label{tab:fixed}
\small
\begin{tabular}{lcc}
\toprule
Method & Distance (m) & Relative completion \\
\midrule
\mseBC & $1430.05\pm580.55$ & $0.83\pm0.26$ \\
\dBC & $1339.85\pm150.60$ & $1.00\pm0.00$ \\
\dTwoBC & $\mathbf{1498.09\pm74.81}$ & $\mathbf{1.00\pm0.00}$ \\
\bottomrule
\end{tabular}
\end{table}

\subsection{Route-free multi-intersection navigation in Town01}
The multi-intersection experiment combines route-choice diversity with long-horizon closed-loop control. At its peak mean-distance checkpoint, \mseBC travels 124.945 UoM on average. Its minimum, maximum, and mean are identical within each evaluation because repeated rollouts execute the same actions and overlap on the same trajectory. This deterministic consistency does not reproduce the route diversity present in the 34-route dataset.

\dBC reaches a lower peak mean of 51.873 UoM and exhibits a wide range of rollout lengths. At some evaluated checkpoints, one rollout travels only 4.501 UoM while another reaches 365.222 UoM. The method therefore generates stochastic trajectory variation, but this variation is not consistently associated with distinct intersection decisions or reliable progress.

\begin{figure}[t!]
\centering
\captionsetup[subfigure]{font=scriptsize}
\begin{subfigure}{0.295\textwidth}
\includegraphics[width=\linewidth,angle=270]{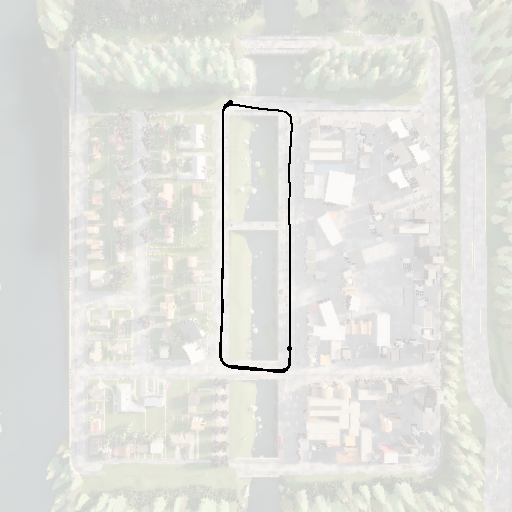}
\caption{Town01 peak: MSE-BC, ep. 60}
\end{subfigure}\hfill
\begin{subfigure}{0.295\textwidth}
\includegraphics[width=\linewidth,angle=270]{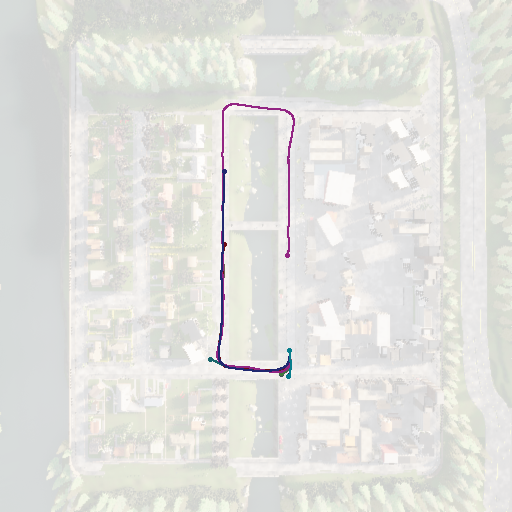}
\caption{Town01 peak: Diffusion-BC, ep. 210}
\end{subfigure}\hfill
\begin{subfigure}{0.295\textwidth}
\includegraphics[width=\linewidth,angle=270]{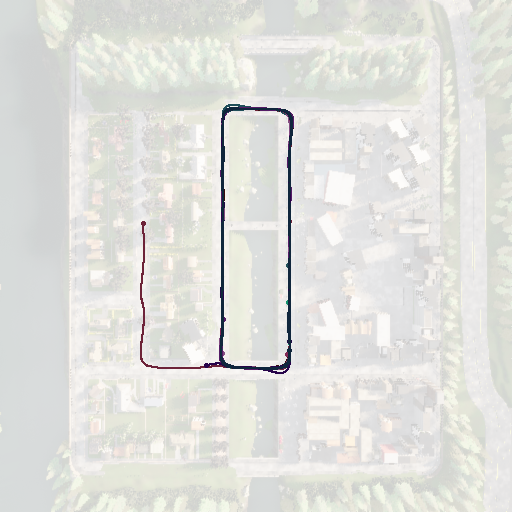}
\caption{Town01 peak: Diffusion-2BC, ep. 330}
\end{subfigure}

\begin{subfigure}{0.295\textwidth}
\includegraphics[width=\linewidth,angle=270]{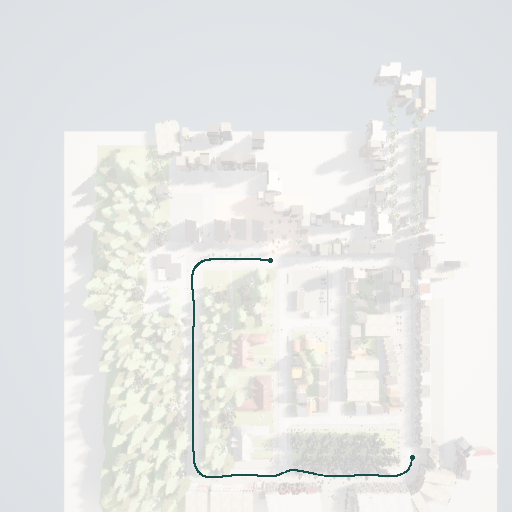}
\caption{Town02 transferred candidate: MSE-BC, ep. 220}
\end{subfigure}\hfill
\begin{subfigure}{0.295\textwidth}
\includegraphics[width=\linewidth,angle=270]{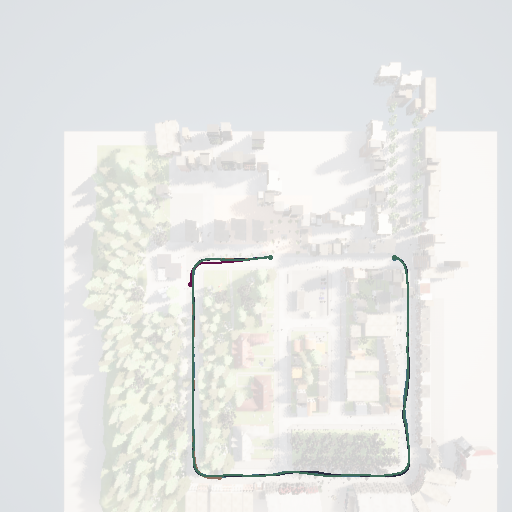}
\caption{Town02 transferred candidate: Diffusion-BC, ep. 230}
\end{subfigure}\hfill
\begin{subfigure}{0.295\textwidth}
\includegraphics[width=\linewidth,angle=270]{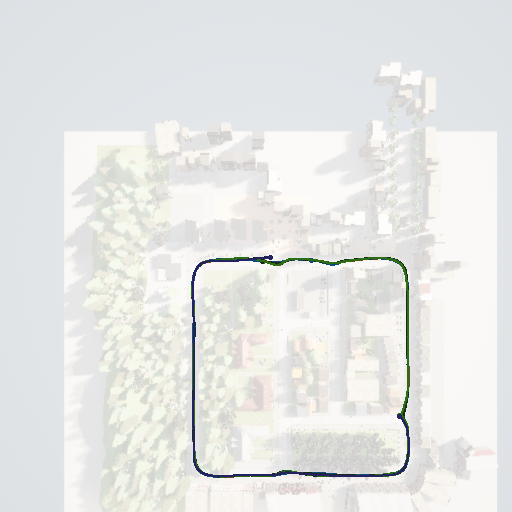}
\caption{Town02 transferred candidate: Diffusion-2BC, ep. 680}
\end{subfigure}

\begin{subfigure}{0.295\textwidth}
\includegraphics[width=\linewidth,angle=270]{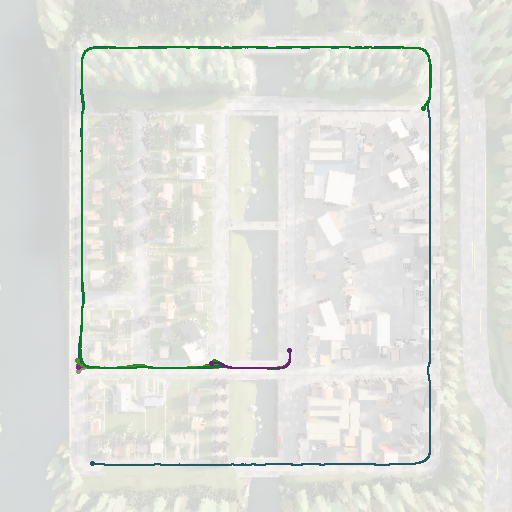}
\caption{Town01 qualitative: Diffusion-2BC, ep. 340}
\end{subfigure}\hfill
\begin{subfigure}{0.295\textwidth}
\includegraphics[width=\linewidth,angle=270]{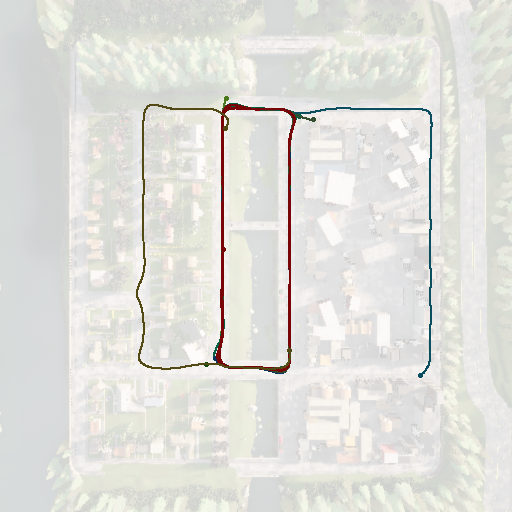}
\caption{Town01 qualitative: Diffusion-2BC, ep. 640}
\end{subfigure}\hfill
\begin{subfigure}{0.295\textwidth}
\includegraphics[width=\linewidth,angle=270]{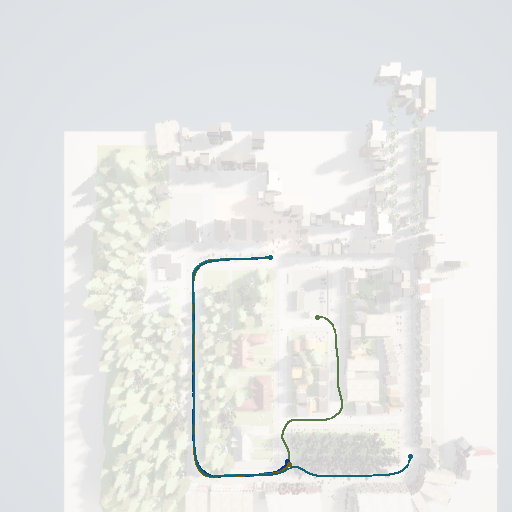}
\caption{Town02 qualitative: Diffusion-2BC, ep. 50}
\end{subfigure}
\caption{Route-free rollouts; each panel overlays ten trajectories from the same initial point. The Town01 cross-method panels use the highest mean-distance checkpoint for each architecture (epochs 60, 210, and 330 for MSE-BC, Diffusion-BC, and Diffusion-2BC). The Town02 panels show epochs 220, 230, and 680 from the fixed set of five checkpoints preselected in Town01; these are the highest observed Town02 means within that transferred set, not checkpoints obtained by a new Town02 search. The last row contains additional Diffusion-2BC checkpoints selected only for qualitative inspection of route diversity. These qualitative panels are not used to compute or select the values in Table~\ref{tab:multi}.}
\label{fig:selectedroutes}
\end{figure}

\dTwoBC attains the highest peak mean, 304.697 UoM, approximately 2.44 times the \mseBC mean and 5.87 times the \dBC mean. Its range at the peak mean-distance checkpoint is 116.741--360.201 UoM. To avoid method-specific hand-selection, the cross-method trajectory comparison in Fig.~\ref{fig:selectedroutes} uses the same objective rule for all architectures: the checkpoint with the highest mean distance. These checkpoints are epoch 60 for \mseBC, epoch 210 for \dBC, and epoch 330 for \dTwoBC. Because peak distance does not necessarily coincide with the clearest route diversity, epochs 340 and 640 of \dTwoBC are shown separately as additional qualitative examples; they are not used for the quantitative comparison in Table~\ref{tab:multi}. The complete Town01 top-checkpoint histograms are provided in Supplementary Fig.~S3.

\subsection{Descriptive cross-map evaluation in Town02}
\begin{figure}[t]
\centering
\begin{subfigure}{0.32\textwidth}
\includegraphics[width=\linewidth]{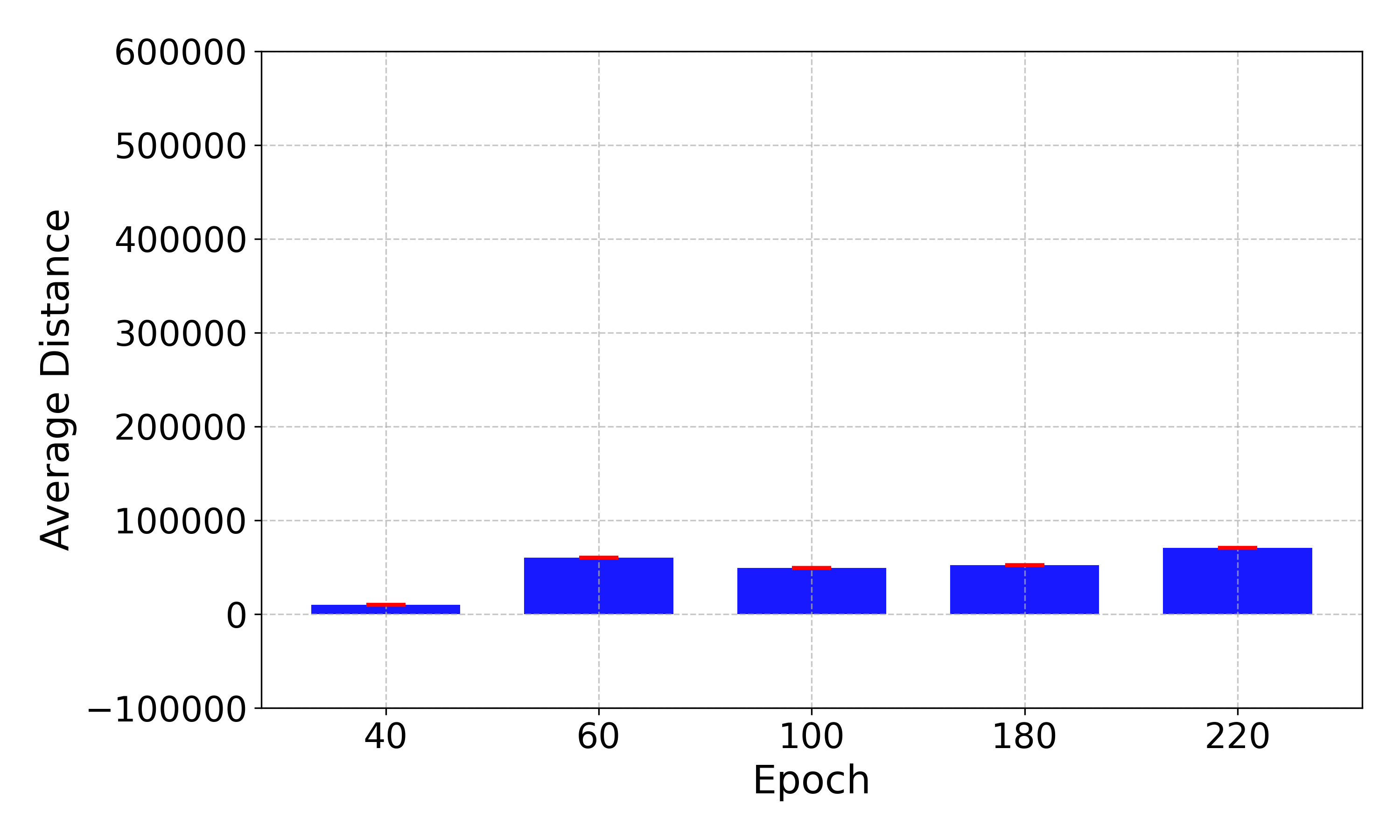}
\caption{MSE-BC}
\end{subfigure}\hfill
\begin{subfigure}{0.32\textwidth}
\includegraphics[width=\linewidth]{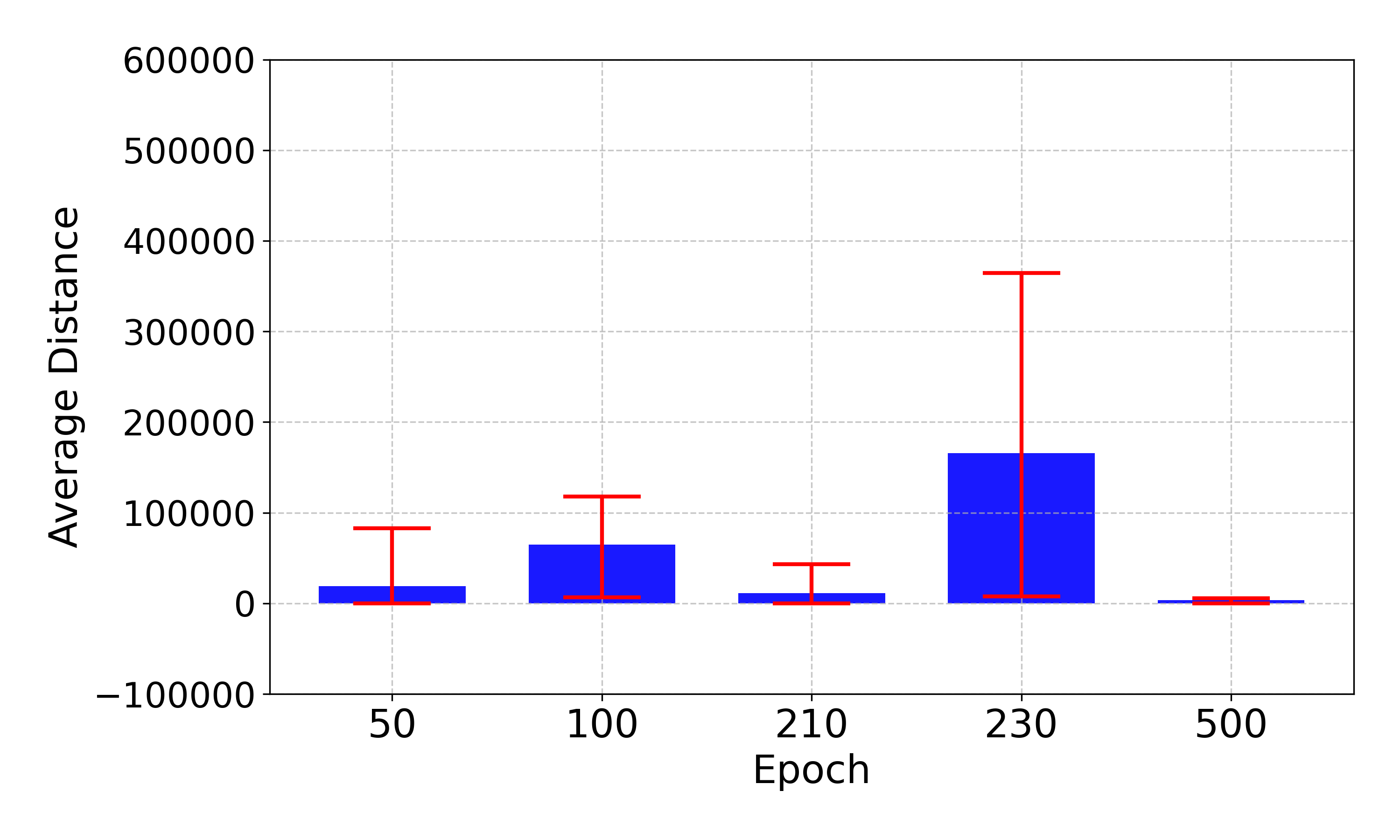}
\caption{Diffusion-BC}
\end{subfigure}\hfill
\begin{subfigure}{0.32\textwidth}
\includegraphics[width=\linewidth]{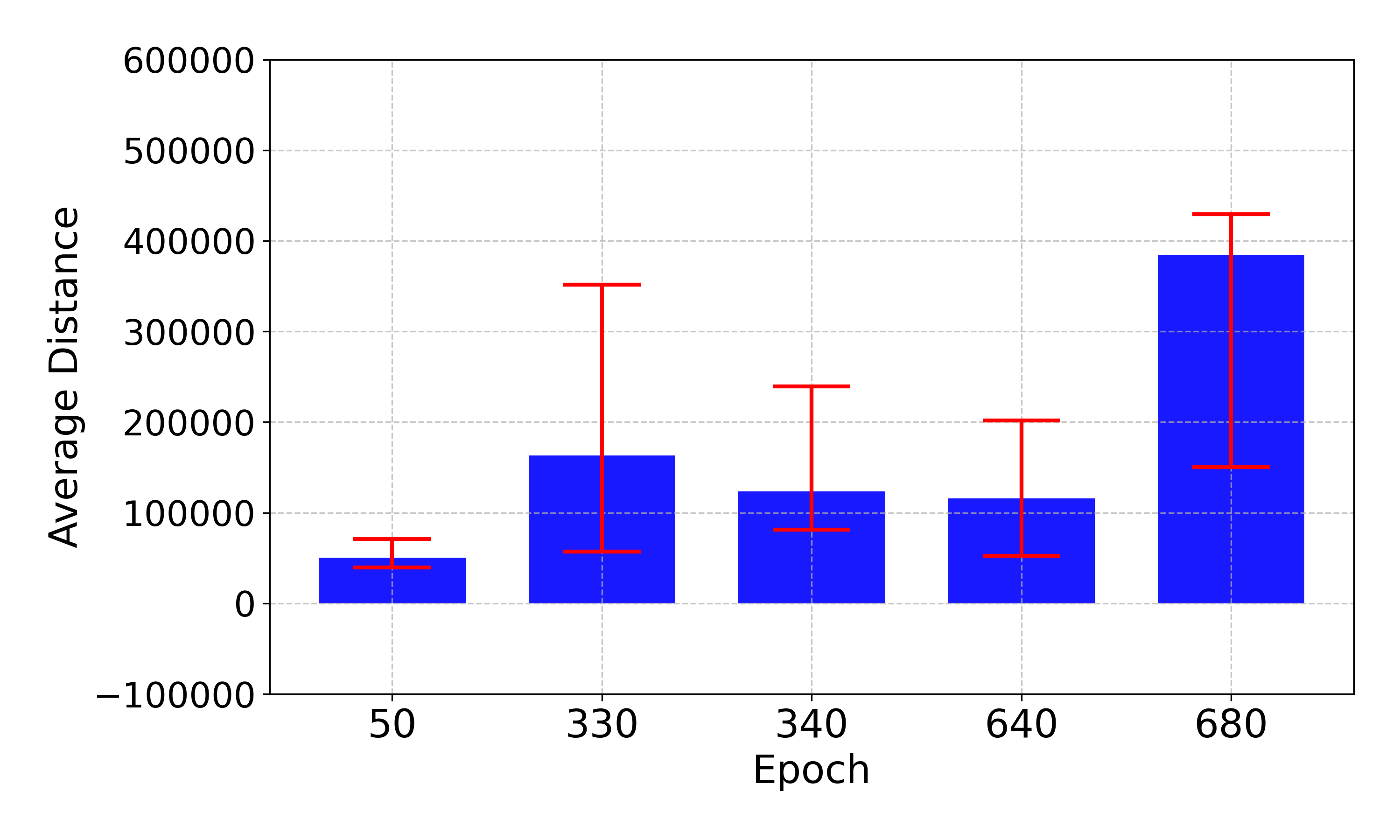}
\caption{Diffusion-2BC}
\end{subfigure}
\caption{Town02 cross-map performance for five checkpoints preselected from Town01 mean-distance performance, using one Town01-trained model per architecture. Bars show mean route-free distance and bounds show the minimum and maximum across repeated rollouts. MSE-BC repeats a single deterministic path, Diffusion-BC has large variability, and Diffusion-2BC achieves the highest peak mean distance}
\label{fig:town02hist}
\end{figure}

The Town02 experiment evaluates policies trained only in Town01, without additional gradient updates, under changed road geometry. For each architecture, the five candidate checkpoints were first selected from Town01 according to mean-distance performance and were then evaluated in Town02 without retraining or a new sweep over training epochs. Fig.~\ref{fig:town02hist} reports Town02 performance for these Town01-preselected checkpoints from one trained instance of each architecture. Among these transferred candidates, \mseBC reaches a highest observed mean of 70.563 UoM and has zero-width min--max bounds, confirming that all repeated rollouts follow the same path. \dBC reaches 165.717 UoM but remains highly inconsistent, with a 7.489--364.232 UoM range. \dTwoBC reaches the highest observed mean, 383.963 UoM, with a 150.107--429.379 UoM range.

Among the five Town01-preselected candidates evaluated in Town02, epochs 220, 230, and 680 yield the highest Town02 mean distances for \mseBC, \dBC, and \dTwoBC, respectively. At epoch 680, the \dTwoBC rollouts predominantly share the same cyclic road structure. Epoch 50 is therefore shown only as an additional qualitative example because it provides the clearest Town02 instance of distinct route choices, including an inner and an outer route, while still demonstrating cross-map operation. This separation prevents a visually selected checkpoint from being presented as the quantitative optimum.

\begin{table}[t]
\centering
\caption{Route-free multi-intersection distance results. All policies are trained in Town01, and Town02 is unseen during training. Town01 candidate checkpoints are selected by mean-distance performance; the Town02 row reports the highest Town02 mean among those five preselected candidates rather than a sweep over all training epochs.}
\label{tab:multi}
\scriptsize
\setlength{\tabcolsep}{2.0pt}
\begin{tabularx}{\textwidth}{>{\raggedright\arraybackslash}p{0.15\textwidth} >{\raggedright\arraybackslash}X >{\centering\arraybackslash}p{0.135\textwidth} >{\centering\arraybackslash}p{0.135\textwidth} >{\centering\arraybackslash}p{0.135\textwidth}}
\toprule
Environment & Metric & \mseBC & \dBC & \dTwoBC \\
\midrule
\multirow{2}{*}{Town01} & Peak mean distance (UoM) & 124.945 & 51.873 & \textbf{304.697} \\
& Reported min--max at peak (UoM) & identical within runs & 0.907--121.469 & \textbf{116.741--360.201} \\
\midrule
\multirow{2}{*}{Town02} & Highest mean in transferred set (UoM) & 70.563 & 165.717 & \textbf{383.963} \\
& Min--max at corresponding candidate (UoM) & identical within runs & 7.489--364.232 & \textbf{150.107--429.379} \\
\bottomrule
\end{tabularx}
\end{table}

\subsection{Computational cost}
In the route-free CARLA experiment, \mseBC requires approximately one minute per ten training epochs and five seconds per 100 inference steps. \dBC and \dTwoBC require approximately ten minutes per ten training epochs and 32 seconds per 100 inference steps. The proposed auxiliary branch does not increase test-time cost relative to \dBC because it is discarded after training. The reported implementation used Python 3.10 with PyTorch 1.13.1 and CUDA 11.7 runtime packages on an NVIDIA GeForce RTX 3060 GPU and an AMD Ryzen 5 5500 CPU.

Deployment-oriented work can be done to increase inference speed and should consider fewer denoising steps, distillation, action chunking, or a faster sampler, but those alternatives were not evaluated and are left for future work.

\section{Conclusions}
This paper presented \dTwoBC, an offline behavior-cloning method that combines a diffusion denoising objective with an auxiliary deterministic regression loss. The auxiliary branch guides the shared visual representation during training and is removed at test time, so inference remains diffusion-based and stochastic.

The staged experiments provide complementary evidence for the proposed training strategy. In the Claw benchmark, diffusion-based objectives substantially reduce error relative to deterministic behavior cloning when valid actions occupy disconnected regions. In route-conditioned CARLA navigation, \dTwoBC{} reaches infraction-free completion earlier and with lower variability than the evaluated \dBC{} baseline. In the route-free experiment, it reaches the highest observed mean distance under the configured evaluation protocol, while selected checkpoints provide qualitative evidence of distinct route choices in Town01 and Town02.

These results support a focused conclusion: within the evaluated datasets and baselines, auxiliary regression improves the closed-loop reliability of diffusion behavior cloning while retaining multimodal prediction in the Claw benchmark and qualitative route variation in CARLA. The experiments are intended to isolate the effect of the auxiliary training signal rather than to establish superiority over all multimodal imitation-learning architectures.

Future work will broaden the comparison to energy-based, transformer-based, and other diffusion-policy formulations; systematically evaluate the weighting schedule; quantify trajectory diversity; reduce diffusion inference cost; and extend the closed-loop evaluation to dynamic traffic, traffic rules, adverse environmental conditions, and operational perception pipelines. Beta-distribution diffusion noise is another possible direction because the driving controls are bounded \cite{Beta_Diffusion}.

\section*{Declarations}
\paragraph{Funding.}
This work was partially funded by CNPq, Brazil (Grant No. 420148/2025-6).

\paragraph{Competing interests.}
The authors have no relevant financial or non-financial interests to disclose.

\paragraph{Data availability.}
The datasets, route definitions, and trained models supporting the findings of this study are available from the corresponding author upon reasonable request.

\paragraph{Code availability.}
The code supporting the findings of this study is available from the corresponding author upon reasonable request.

\paragraph{Author contributions.}
Bruno Maciel Machado: Methodology, Software, Investigation, Data Curation, Visualization, and Writing---Original Draft. Eric Aislan Antonelo: Conceptualization, Methodology, Formal Analysis, Supervision, Validation, Project Administration, Writing---Original Draft, and Writing---Review and Editing.

\clearpage
\section*{Supplementary Material}
\setcounter{figure}{0}
\setcounter{table}{0}

\section*{Supplementary Note S1: Preliminary CarRacing experiment}
The main text cites the CarRacing experiment as preliminary empirical motivation for combining diffusion-based action generation with a direct regression signal. The experiment predates \dTwoBC{} and compares only deterministic \mseBC{} and \dBC{}; it is therefore not used as validation of the proposed method. This note records the protocol, reproduces the previously published reward distributions and aggregate statistics for completeness, and adds cumulative-reward visualizations that were not included in the preliminary conference article~\cite{Machado2025CROS}.

\paragraph{Environment and dataset.}
CarRacing-v2 provides $96\times96$ RGB observations and a continuous action vector containing steering, acceleration, and braking. Each observation was converted to grayscale, stacked with the three preceding frames, and normalized, resulting in a $96\times96\times4$ input. The fixed offline dataset contains approximately 15,000 observation--action pairs collected from 20 manually driven expert episodes on randomly generated tracks. Every demonstration obtained a reward above 900; the expert mean was $930.34\pm7.46$.

\paragraph{Training and evaluation.}
Both policies used the same convolutional feature extractor and were trained for 80 epochs with learning rate $10^{-4}$, batch size 64, and 512 hidden feature units. \dBC{} used 50 diffusion timesteps, a linear $\beta$ schedule from $10^{-4}$ to $2\times10^{-2}$, and a 128-dimensional embedding. Two independently trained models of each architecture were evaluated on 100 consecutive randomly generated tracks per model. A reward above 900 indicates expert-level completion under the environment's progress-and-time reward.

\begin{center}
\small
\textbf{Supplementary Table S1.} Aggregate CarRacing performance across two independently trained models, each evaluated on 100 randomly generated tracks. These values were reported in the preliminary conference article.\par\medskip
\begin{tabular}{lccc}
\toprule
Method & Mean reward & Standard deviation & Episodes above 900 \\
\midrule
\mseBC{} & 638.79 & 105.20 & 0\% \\
\dBC{} & \textbf{861.35} & \textbf{48.07} & \textbf{41\%} \\
\bottomrule
\end{tabular}
\end{center}

\begin{figure}[t]
\centering
\begin{subfigure}{0.48\textwidth}
\includegraphics[width=\linewidth]{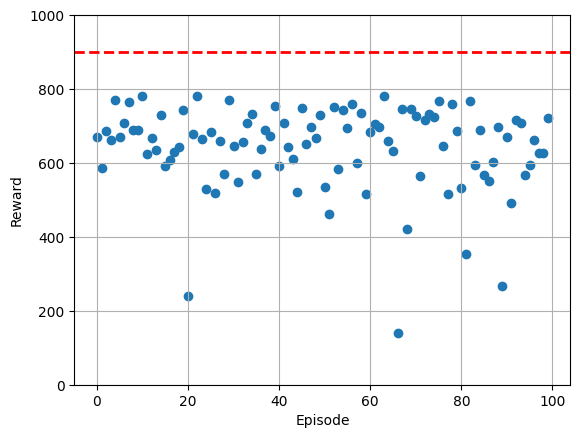}
\caption{\mseBC{}: $642.73\pm113.10$.}
\end{subfigure}\hfill
\begin{subfigure}{0.48\textwidth}
\includegraphics[width=\linewidth]{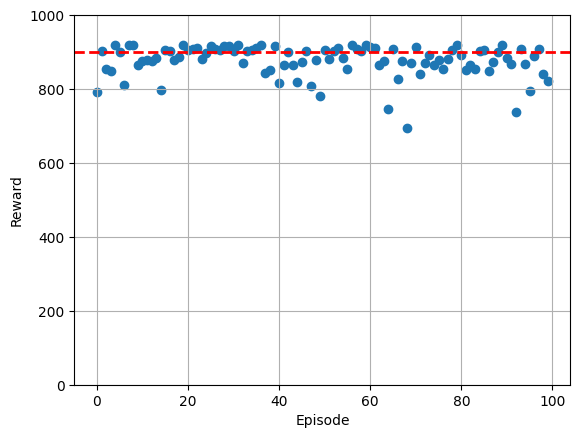}
\caption{\dBC{}: $876.39\pm42.96$.}
\end{subfigure}
\caption*{Supplementary Fig. S1. Reward distributions over 100 randomly generated evaluation tracks for one representative trained model of each architecture. The dashed red line marks reward 900. These distributions were previously reported in the conference article and are reproduced here to make the supplementary account self-contained.}
\end{figure}

\begin{figure}[t]
\centering
\begin{subfigure}{0.49\textwidth}
\includegraphics[width=\linewidth]{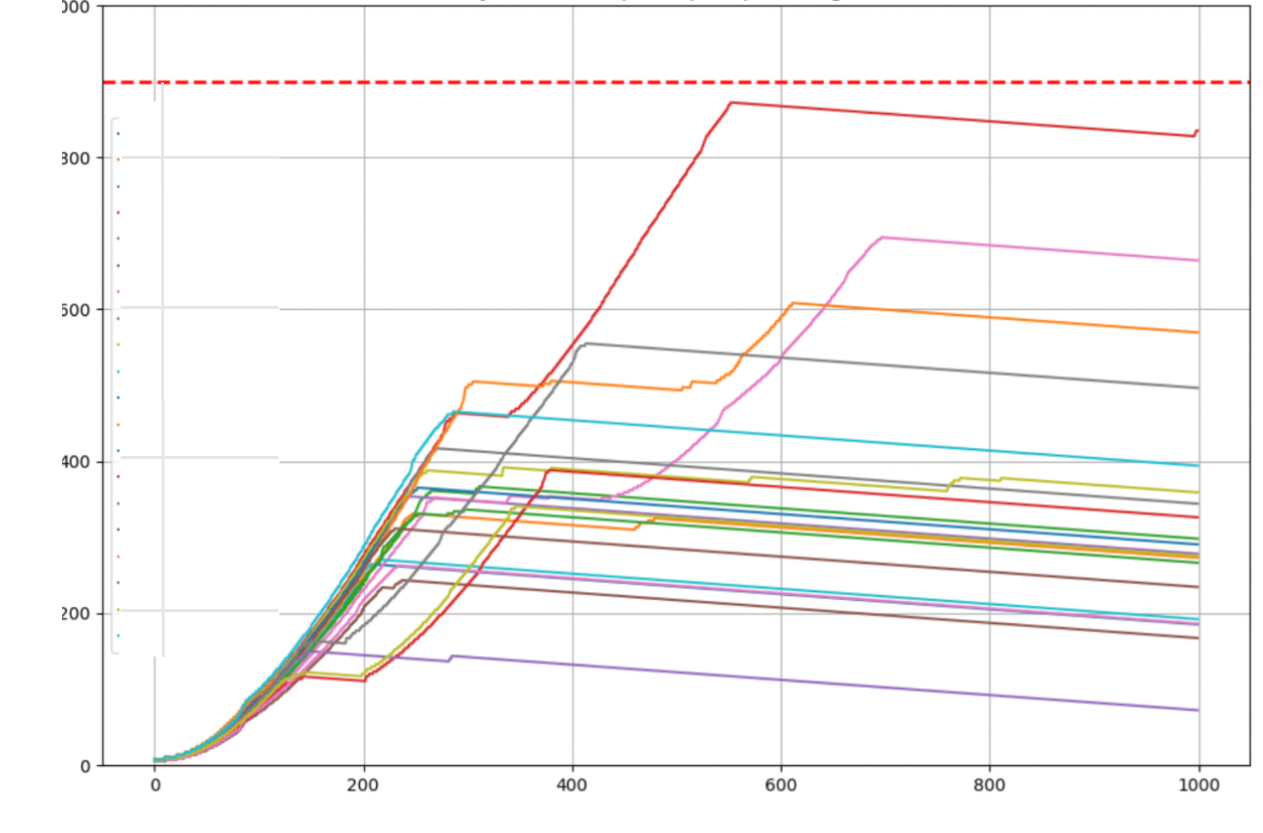}
\caption{\mseBC{}.}
\end{subfigure}\hfill
\begin{subfigure}{0.49\textwidth}
\includegraphics[width=\linewidth]{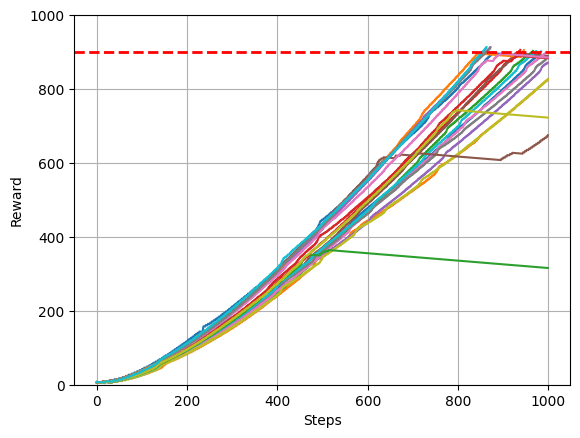}
\caption{\dBC{}.}
\end{subfigure}
\caption*{Supplementary Fig. S2. Cumulative reward over environment steps for 20 representative evaluation episodes from one trained model of each architecture. These curves were not included in the preliminary conference article. The \mseBC{} curves frequently plateau after missed track tiles, whereas most \dBC{} curves continue toward successful completion. The plots provide qualitative context for the aggregate statistics and are not evidence about \dTwoBC{}.}
\end{figure}

\clearpage
\section*{Supplementary Table S2: Layer-by-layer policy architectures}
The table below expands the CARLA architecture summary in the main text; the preliminary CarRacing configuration is documented separately in Supplementary Note S1. In CARLA, the four-frame $192\times192$ input produces a 4,608-dimensional flattened visual representation and the action dimension is $d_a=2$.

\begin{table}[h]
\centering
\caption*{Supplementary Table S2. Detailed CARLA neural-network components used by MSE-BC, Diffusion-BC, and Diffusion-2BC.}
\scriptsize
\begin{tabularx}{\textwidth}{>{\raggedright\arraybackslash}p{0.20\textwidth}>{\raggedright\arraybackslash}p{0.52\textwidth}X}
\toprule
Component & Layer sequence and dimensions & Used by \\
\midrule
Residual block 1 & Two Conv2D layers, kernel $3\times3$, stride 1, padding 1, 64 output channels; BatchNorm and GELU after each convolution; residual sum scaled by $1/\sqrt{2}$; MaxPool2D $2\times2$ & All policies \\
Residual block 2 & Two Conv2D layers with the same kernel, stride, padding, normalization, activation, and residual scaling; 128 output channels; MaxPool2D $2\times2$ & All policies \\
Visual projection & AvgPool2D $8\times8$, followed by flattening; resulting dimension depends on observation resolution & All policies \\
Deterministic policy head & Flattened visual vector $\rightarrow2304\rightarrow1024\rightarrow64\rightarrow d_a$ with ReLU after the hidden linear layers & \mseBC{} \\
Observation embedding & 4,608-dimensional flattened visual vector $\rightarrow128\rightarrow128$ with LeakyReLU between the linear layers for CARLA & \dBC{}, \dTwoBC{} \\
Noisy-action embedding & $d_a\rightarrow128\rightarrow128$ with LeakyReLU between the linear layers & \dBC{}, \dTwoBC{} \\
Timestep embedding & Scalar normalized timestep processed by a two-layer TimeSiren embedding with a sine activation, producing 128 features & \dBC{}, \dTwoBC{} \\
Transformer tokens & Observation, noisy-action, and timestep embeddings are separately projected from 128 to 64 features and supplied as three tokens with positional embeddings & \dBC{}, \dTwoBC{} \\
Transformer denoiser & Four encoder blocks; 16 attention heads; internal Q/K/V dimension $64\times16=1024$; feed-forward path $64\rightarrow256\rightarrow64$ with GELU; residual scaling and BatchNorm; no Transformer dropout & \dBC{}, \dTwoBC{} \\
Noise projection & Flatten three 64-dimensional output tokens and project $192\rightarrow d_a$ to predict the action noise & \dBC{}, \dTwoBC{} \\
Auxiliary regression head & Flattened visual vector $\rightarrow2048\rightarrow512\rightarrow128\rightarrow32\rightarrow d_a$; BatchNorm, ReLU, and dropout $p=0.3$ after each hidden linear layer & \dTwoBC{} training only \\
Inference & Direct deterministic action for \mseBC{}; iterative reverse diffusion for \dBC{} and \dTwoBC{}; auxiliary regression head omitted for \dTwoBC{} & Method-dependent \\
\bottomrule
\end{tabularx}
\end{table}
\FloatBarrier
\clearpage

\section*{Supplementary Table S3: CARLA experimental hyperparameters}
The main text reports experiment-specific settings in the methodology. Supplementary Table S3 consolidates the principal training and diffusion hyperparameters used in the CARLA experiments for convenient replication.

\begin{table}[h]
\centering
\caption*{Supplementary Table S3. Main training and diffusion hyperparameters for the CARLA experiments.}
\small
\begin{tabularx}{\textwidth}{>{\raggedright\arraybackslash}p{0.25\textwidth}XX}
\toprule
Parameter & Route-conditioned CARLA & Route-free CARLA \\
\midrule
Learning rate & $10^{-4}$ & $10^{-4}$ \\
Learning-rate schedule & Fixed & Fixed \\
Feature hidden units & 128 & 128 \\
Batch size & 32 & 32 \\
Optimizer & Adam & Adam \\
Weight decay & 0 & 0 \\
Explicit training seed & Not set & Not set \\
Training epochs & 300 for the primary comparison; \dTwoBC{} monitored to 750 & 700 \\
Diffusion timesteps $T$ & 20 & 20 \\
$\beta$ limits & $10^{-4}$ to $2\times10^{-2}$ & $10^{-4}$ to $2\times10^{-2}$ \\
$\beta$ schedule & Linear & Linear \\
Input embedding / denoiser & 128 / Transformer & 128 / Transformer \\
Context dropout & 0 & 0 \\
Guidance weight & 0 & 0 \\
\dTwoBC{} $\alpha$ & Fixed at 0.3 & Exponential: $\alpha_e=\exp(-\lambda e)$, $\lambda=\ln(100)/E$; updated once per epoch \\
Evaluation frequency & Every 50 epochs & Every 10 epochs \\
\bottomrule
\end{tabularx}
\end{table}
\FloatBarrier

\medskip
\section*{Supplementary Fig. S3: Town01 top-checkpoint performance}
The main text prioritizes the Town02 cross-map histograms because performance under changed road geometry is more informative than training-map performance. For completeness, Supplementary Fig. S3 reports the corresponding Town01 training-map histograms for one trained model per architecture.

\begin{figure}[h]
\centering
\begin{subfigure}{0.32\textwidth}
\includegraphics[width=\linewidth]{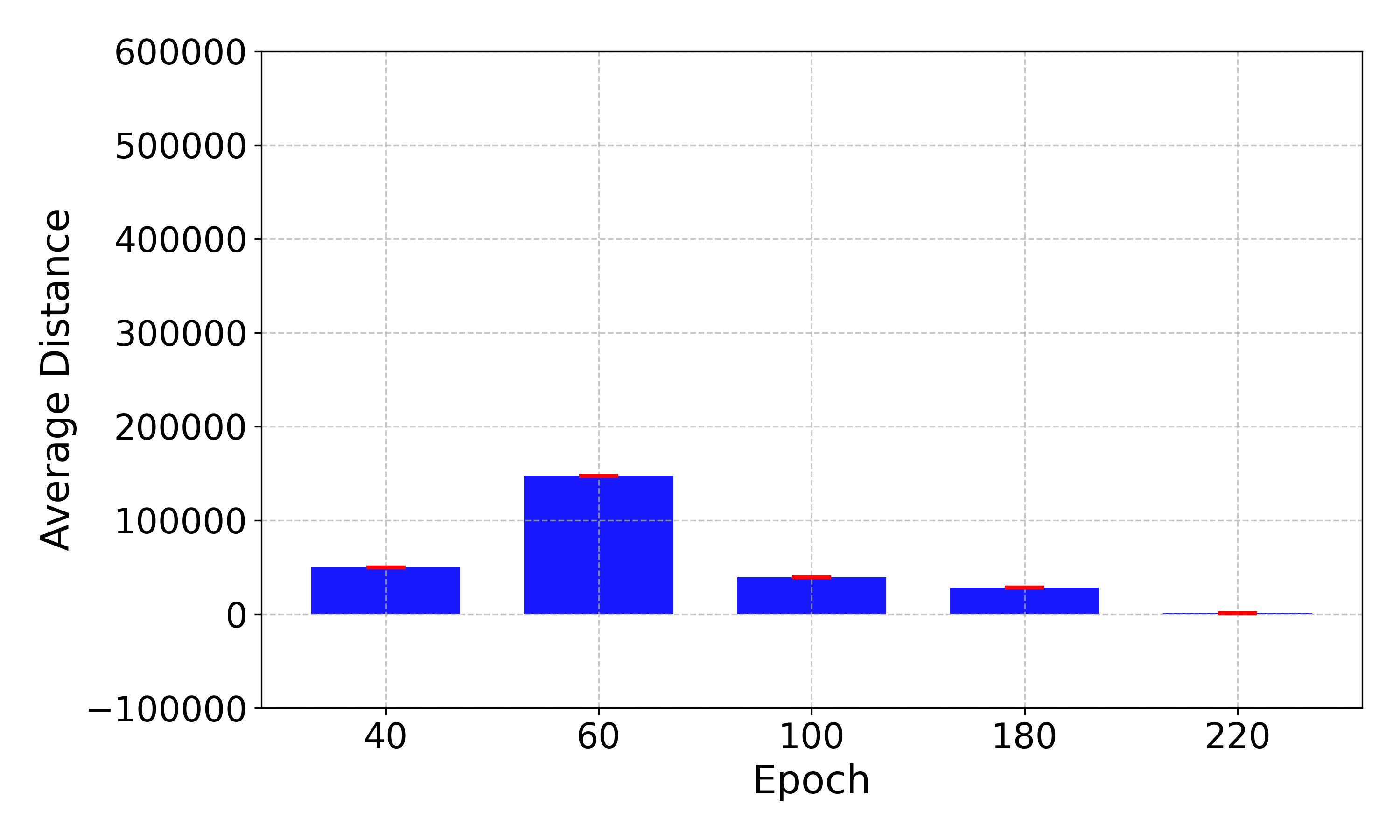}
\caption{MSE-BC.}
\end{subfigure}\hfill
\begin{subfigure}{0.32\textwidth}
\includegraphics[width=\linewidth]{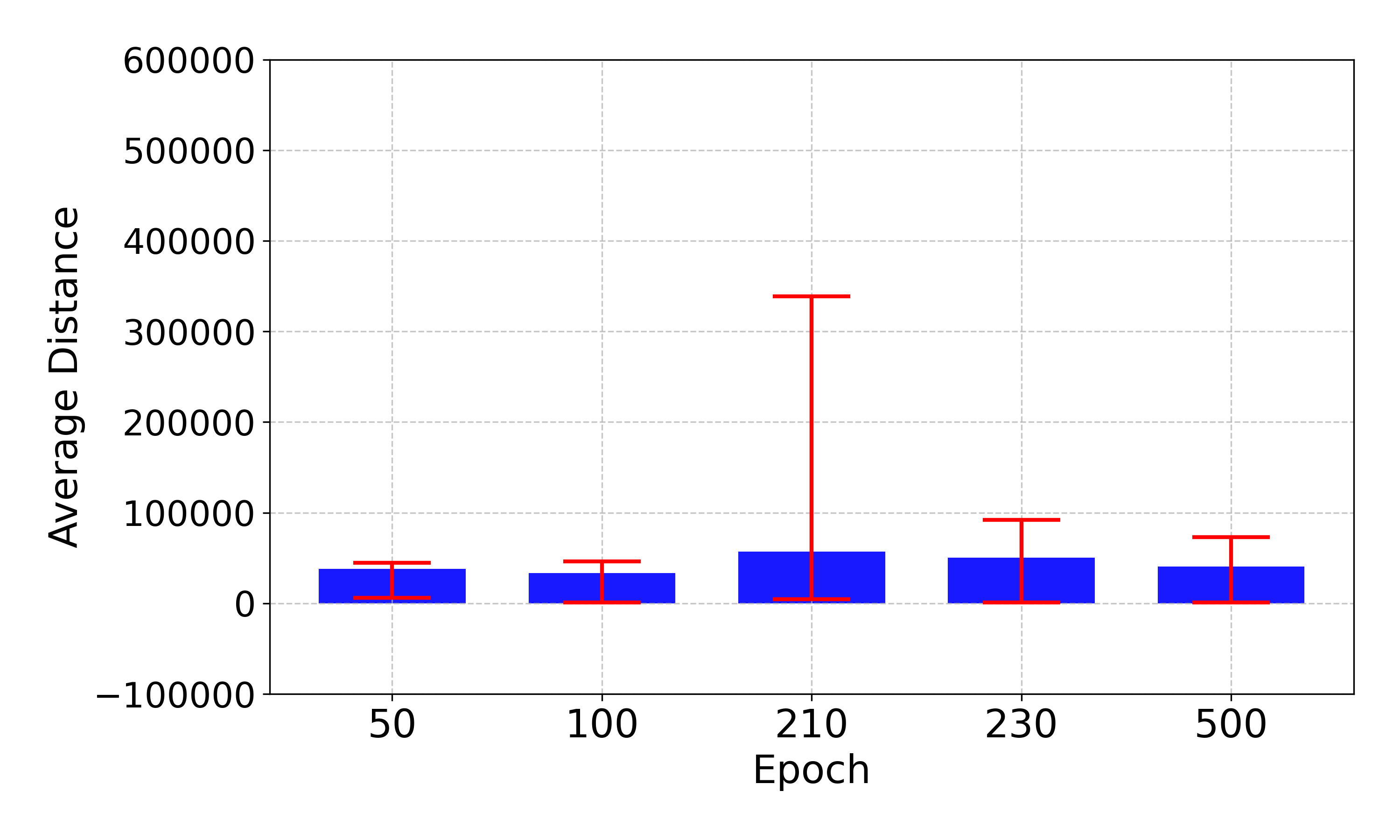}
\caption{Diffusion-BC.}
\end{subfigure}\hfill
\begin{subfigure}{0.32\textwidth}
\includegraphics[width=\linewidth]{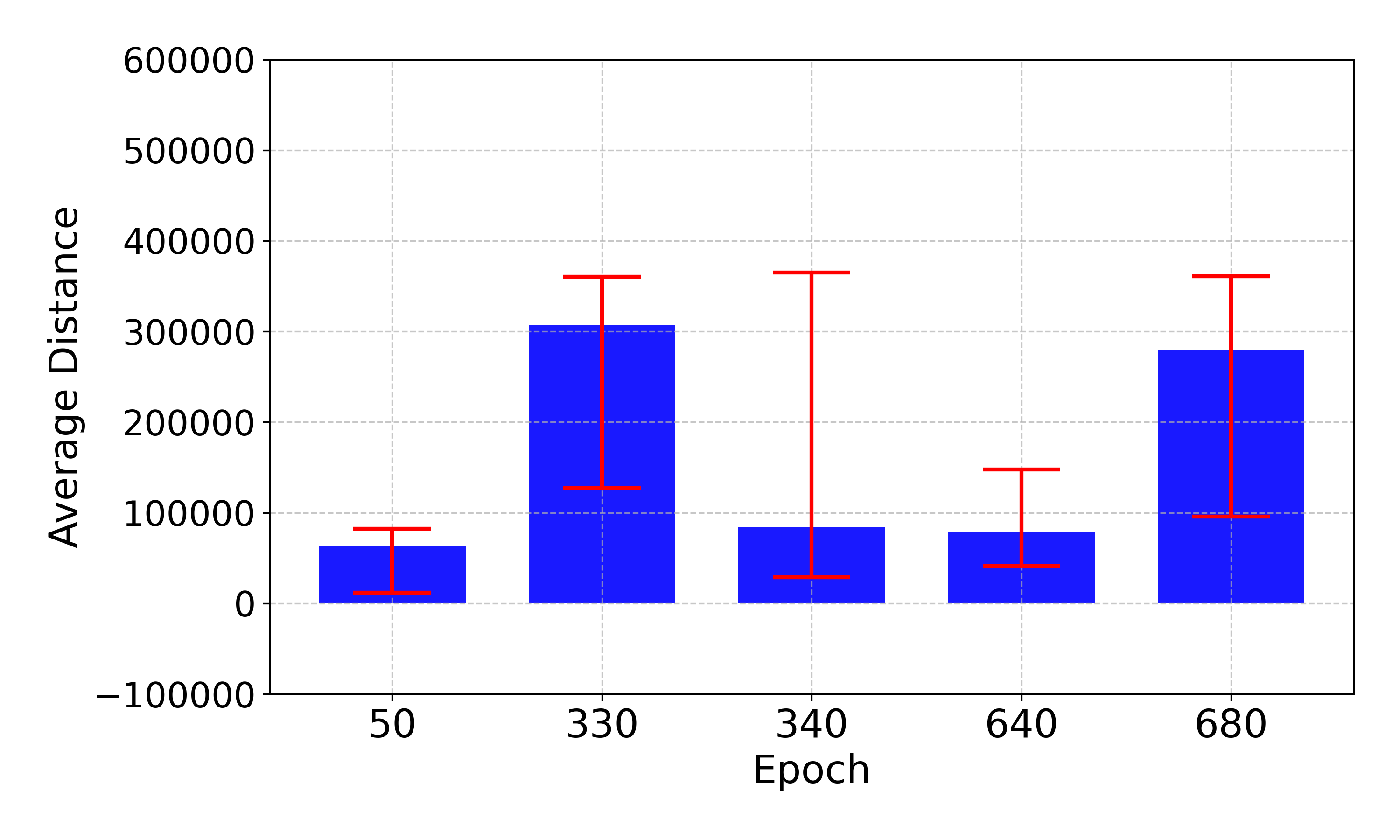}
\caption{Diffusion-2BC.}
\end{subfigure}
\caption*{Supplementary Fig. S3. Town01 route-free performance for the five highest mean-distance checkpoints of one trained model per architecture. Bars show mean distance and bounds show the minimum and maximum across repeated rollouts.}
\end{figure}

\clearpage
\bibliographystyle{unsrtnat}
\bibliography{references}
\end{document}